\PassOptionsToPackage{unicode}{hyperref}
\PassOptionsToPackage{hyphens}{url}
\documentclass[
  12pt,
]{article}
\usepackage{xcolor}
\usepackage[margin=1in]{geometry}
\usepackage{amsmath,amssymb}
\usepackage{iftex}
\ifPDFTeX
  \usepackage[T1]{fontenc}
  \usepackage[utf8]{inputenc}
  \usepackage{textcomp} 
\else 
  \usepackage{unicode-math} 
  \defaultfontfeatures{Scale=MatchLowercase}
  \defaultfontfeatures[\rmfamily]{Ligatures=TeX,Scale=1}
\fi
\usepackage{lmodern}
\ifPDFTeX\else
\fi
\IfFileExists{upquote.sty}{\usepackage{upquote}}{}
\IfFileExists{microtype.sty}{
  \usepackage[]{microtype}
  \UseMicrotypeSet[protrusion]{basicmath} 
}{}
\makeatletter
\@ifundefined{KOMAClassName}{
  \IfFileExists{parskip.sty}{%
    \usepackage{parskip}
  }{
    \setlength{\parindent}{0pt}
    \setlength{\parskip}{6pt plus 2pt minus 1pt}}
}{
  \KOMAoptions{parskip=half}}
\makeatother
\usepackage{longtable,booktabs,array}
\usepackage{calc} 
\usepackage{etoolbox}
\makeatletter
\patchcmd\longtable{\par}{\if@noskipsec\mbox{}\fi\par}{}{}
\makeatother
\IfFileExists{footnotehyper.sty}{\usepackage{footnotehyper}}{\usepackage{footnote}}
\makesavenoteenv{longtable}
\usepackage{graphicx}
\makeatletter
\newsavebox\pandoc@box
\newcommand*\pandocbounded[1]{
  \sbox\pandoc@box{#1}%
  \Gscale@div\@tempa{\textheight}{\dimexpr\ht\pandoc@box+\dp\pandoc@box\relax}%
  \Gscale@div\@tempb{\linewidth}{\wd\pandoc@box}%
  \ifdim\@tempb\p@<\@tempa\p@\let\@tempa\@tempb\fi
  \ifdim\@tempa\p@<\p@\scalebox{\@tempa}{\usebox\pandoc@box}%
  \else\usebox{\pandoc@box}%
  \fi%
}
\def\fps@figure{htbp}
\makeatother
\NewDocumentCommand\citeproctext{}{}

\makeatletter
 \let\@cite@ofmt\@firstofone
 \def\@biblabel#1{}
 \def\@cite#1#2{{#1\if@tempswa , #2\fi}}
\makeatother
\newlength{\cslhangindent}
\newlength{\csllabelwidth}
\newenvironment{CSLReferences}[2] 
 {\begin{list}{}{%
  \setlength{\itemindent}{0pt}
  \setlength{\leftmargin}{0pt}
  \setlength{\parsep}{0pt}
  \ifodd #1
   \setlength{\leftmargin}{\cslhangindent}
   \setlength{\itemindent}{-1\cslhangindent}
  \fi
  \setlength{\itemsep}{#2\baselineskip}}}
 {\end{list}}
\usepackage{calc}

\providecommand{\tightlist}{%
  \setlength{\itemsep}{0pt}\setlength{\parskip}{0pt}}
\usepackage{booktabs}
\usepackage{graphicx}
\usepackage{float}
\usepackage{amsmath}
\usepackage{amssymb}
\usepackage{bookmark}
\IfFileExists{xurl.sty}{\usepackage{xurl}}{} 
\makeatletter
\@ifundefined{xmpquote}{}{}
\makeatother
\hypersetup{
  pdftitle={Multicalibration for Unbiased Model-Based Prevalence Estimation},
  pdfauthor={Fridolin Linder1; Thomas Leeper1; Daniel Haimovich1; Niek Tax1; Lorenzo Perini1; Milan Vojnovic1,2},
  hidelinks,
  pdfcreator={LaTeX via pandoc}}

\title{Multicalibration for Unbiased Model-Based Prevalence Estimation}
\author{Fridolin Linder\textsuperscript{1} \and Thomas
Leeper\textsuperscript{1} \and Daniel
Haimovich\textsuperscript{1} \and Niek
Tax\textsuperscript{1} \and Lorenzo Perini\textsuperscript{1} \and Milan
Vojnovic\textsuperscript{1,2}}
\date{}

\begin{document}
\maketitle

\textsuperscript{1}Meta Platforms Inc., \textsuperscript{2}The London
School of Economics and Political Science

Corresponding author: Fridolin Linder (fridolin.linder@gmail.com)

\subsection*{Abstract}\label{abstract}
\addcontentsline{toc}{subsection}{Abstract}

Political scientists increasingly use classifiers, including large
language models (LLMs), to estimate how prevalent a concept is in a
population. Standard validation reports discriminative metrics such as
accuracy or the area under the ROC curve (AUC) on a labeled sample,
which certifies neither unbiased prevalence estimates in the validated
population nor their transport to new ones. Existing corrections, rarely
used in practice, either assume label shift or must be re-estimated for
every target. Building on Kim et al. (2022), we show that a
\emph{multicalibrated} classifier, one calibrated within feature-defined
subgroups rather than only on average, needs to be fitted only once and
yields unbiased prevalence estimates under any covariate shift that the
calibrated groups can express, within their support. In a simulation and
two applications, an LLM coding policy topics across four countries and
languages and a logistic regression predicting employment from survey
data, multicalibration keeps prevalence error below half a percentage
point for shifts within the calibrated features, where standard
corrections miss by up to 19 points, and degrades only modestly beyond
them.

\textbf{Keywords:} prevalence estimation, multicalibration, covariate
shift, quantification, large language models, measurement validity

\section{Introduction}\label{introduction}

Machine learning classifiers, and increasingly large language models
(LLMs), have become standard tools for measuring political concepts from
text, images and administrative records (Benoit et al. 2026; Overos et
al. 2024; Mellon et al. 2024; Ornstein et al. 2025; Torres and Cantú
2022; Imai and Khanna 2016). The quantity of interest is often not the
label on any single unit but an \emph{aggregate} count or prevalence
(Hopkins and King 2010; {González et al.} 2017; Esuli et al. 2023).

Such classifiers are typically validated by how well they classify units
or agree with human coders, using accuracy, \(F_1\), AUC or intercoder
reliability (e.g., Birkenmaier et al. 2024; Törnberg 2024; Knox et al.
2022; Halterman and Keith 2026; Grimmer and Stewart 2013; Barberá et al.
2021; Gilardi et al. 2023; Burnham 2025). For prevalence estimation this
is insufficient, for two reasons. First, a classifier can rank units
almost perfectly yet misstate the share of positives, because its false
positives and false negatives need not cancel. Second, even a correction
estimated on the labeled sample is tied to that sample's population.
When the instrument is applied to a new target, such as another
sub-corpus or a later period, the mix of units can change (covariate
shift). If the target holds more of the cases the classifier misjudges,
errors that canceled in the validation sample no longer do: validity
established in one context need not hold in another (Adcock and Collier
2001; Torres and Cantú 2022; Decter-Frain 2022).

In practice, most applied work simply counts the classifier's labels,
known as Classify \& Count (Esuli et al. 2023). Corrections exist, but
each addresses only part of the problem, and none yields a
\emph{reusable} instrument. Adjusted Classify \& Count (Rogan and Gladen
1978; Saerens et al. 2002) and the classifier-free ReadMe family
(Hopkins and King 2010; Jerzak et al. 2023) carry class-conditional
quantities, such as error rates or word profiles within each category,
from the labeled sample to the target. This is justified under label
shift, where only category frequencies change (Tasche 2023), but
covariate shift generally changes those quantities (Tasche 2022).
Design-based supervised learning (Egami et al. 2023) and
prediction-powered inference (Angelopoulos et al. 2023) require gold
labels drawn from the population being analyzed or, for the latter, a
label shift or known covariate-shift weights linking the two. Importance
weighting (Shimodaira 2000; Sugiyama et al. 2007; Tasche 2023) shares
the covariate-shift assumption but must be re-estimated for each target,
and it sets the classifier aside: its estimate is a reweighted average
of hand codes, with nothing to reweight where the target lies outside
the source's support. Work on learned proxies (Knox et al. 2022), LLM
annotators (Ziems et al. 2024) and LLMs as substitutes for survey
respondents (Argyle et al. 2023; Bisbee et al. 2024) raises the same
worry that imperfect measures distort inference; Weidmann et al. (2026)
find that LLM coders deviate from experts in systematic directions.

Averaging a classifier's scores gives an unbiased prevalence estimate if
the classifier is \emph{calibrated}, that is, if its scores match
empirical frequencies in the population at hand. Wu and Resnick (2024),
for example, calibrate scores on a labeled sample and extrapolate to the
target by assuming a stable calibration curve. But when the classifier's
errors vary with the features, a feature shift that the score does not
capture changes what a given score means, and the calibration curve can
move (Tasche 2022). Building on Kim et al. (2022), we argue that
\emph{multicalibration}, calibration within each of a rich class of
feature-defined subgroups (Hébert-Johnson et al. 2018), addresses this
problem: a model multicalibrated once on a labeled source sample returns
unbiased prevalence estimates on any target whose shift the features
capture and that lies within their support.

\section{Prevalence Bias Reduction through
Multicalibration}\label{prevalence-bias-reduction-through-multicalibration}

Consider a binary outcome \(Y \in \{0,1\}\), features \(X\), and a
predictor \(h(X)\in[0,1]\). The estimand is the target prevalence
\(\pi^*=P^*(Y=1)\), estimated by averaging predictions over unlabeled
target observations. We assume \emph{covariate shift}: \(P(X)\) may
differ between source and target while \(P(Y\mid X)\) is stable, in
contrast to \emph{label shift}, which holds \(P(X\mid Y)\) fixed
(Moreno-Torres et al. 2012). We also require overlap: any feature region
that occurs in the target also occurs in the source. Under concept
drift, where \(P(Y\mid X)\) changes, target prevalence cannot generally
be recovered without target labels.

The prevalence error of a predictor is its mean signed error,
\(\mathbb{E}[h(X)-Y]\). Discriminative metrics do not constrain this
error. For example, AUC depends only on how the scores rank units, so it
is unchanged by any monotone transformation, yet such transformations
can move the average score almost anywhere in \((0,1)\). Accuracy and
\(F_1\) count errors without their sign, whereas the prevalence error of
Classify \& Count is the difference between its false-positive and
false-negative shares. For instance, a classifier with 90\% accuracy can
misstate prevalence by anything from zero, if its errors balance, to ten
percentage points, if they all go one way.

Calibration can close this gap. If \(\mathbb{E}[Y\mid h(X)=p]=p\) for
every prediction value \(p\), then \(\mathbb{E}[h(X)]=\mathbb{E}[Y]\).
But calibration holds in a specific population. Partition that
population into feature-defined groups \(G\) with weights \(w_G\), and
let \(\epsilon_G=\mathbb{E}[h(X)-Y\mid X\in G]\) be the mean signed
error in group \(G\). Calibration ensures only that these errors cancel
under the source weights, \(\sum_G w_G\epsilon_G=0\). If covariate shift
reweights the groups, the errors stay fixed but the weights change, and
the prevalence error becomes
\(\mathbb{E}^*[h(X)]-\pi^*=\sum_G (w_G^*-w_G)\,\epsilon_G\). A
classifier that discriminates well and is calibrated overall can thus
still err systematically within groups, and these errors become
prevalence bias once the group mix changes.

Multicalibration removes these group errors rather than relying on their
cancellation. Let \(\mathcal{G}\) contain the groups that can be formed
from the calibrated features, including their intersections. A predictor
\(f\) is multicalibrated with respect to \(\mathcal{G}\) if
\(\mathbb{E}[Y\mid f(X)=v,\,X\in G]=v\) for every \(G\in\mathcal{G}\)
and prediction value \(v\) (Hébert-Johnson et al. 2018). Averaging over
\(v\) shows that \(f\) has zero mean signed error within every group in
\(\mathcal{G}\). A target shift is captured by \(\mathcal{G}\) if it can
be expressed as a reweighting of these groups. Setting \(\epsilon_G=0\)
in the decomposition above then gives \(\mathbb{E}^*[f(X)]=\pi^*\) for
every captured target within the source's support. This is the
``universal adaptability'' guarantee of Kim et al. (2022), who show that
such an estimate is as accurate as inverse propensity weighting (IPW),
up to the multicalibration error, under any propensity model whose odds
the group class can
express.\footnote{The weaker property of \emph{multi-accuracy}, zero mean error within each group, already suffices. SI Section S1.8 explains why we nonetheless recommend multicalibration.}
The same predictor serves every such target, without per-target
correction or target labels.

The guarantee is exact only when the target-to-source density ratio lies
in the span of the group indicators; otherwise it is approximate, with
error governed by how well the groups approximate that ratio. With
finite calibration data, multicalibration itself also holds only
approximately. Whether covariate shift and this representation condition
are plausible is a substantive judgment about the application. We obtain
the multicalibrated predictor with MCGrad (Tax et al. 2026), a post-hoc
gradient-boosting multicalibrator.

\section{Simulation}\label{simulation}

We illustrate the mechanism with a simulation. Each observation has a
binary feature \(X\) and a continuous signal \(U\sim N(0,1)\), and
\(P(Y=1\mid X,U)=\sigma(a_X+1.5\,U)\) in the calibration sample and all
targets, so only \(P(X)\) shifts. The classifier's score adds a logit
offset for \(X=0\) observations only, so it overstates their probability
while still discriminating well (AUC \(\approx0.89\)). Each estimator is
fit on 10,000 labeled observations at \(P(X=0)=0.5\) and applied to
unlabeled targets with \(P(X=0)\) from 0.01 to 0.99, over 50
repetitions. We compare MCGrad with Classify \& Count, the Rogan-Gladen
adjustment, and global recalibration by isotonic regression (details in
SI Section S4).

\includegraphics[width=0.88\linewidth,height=\textheight,keepaspectratio]{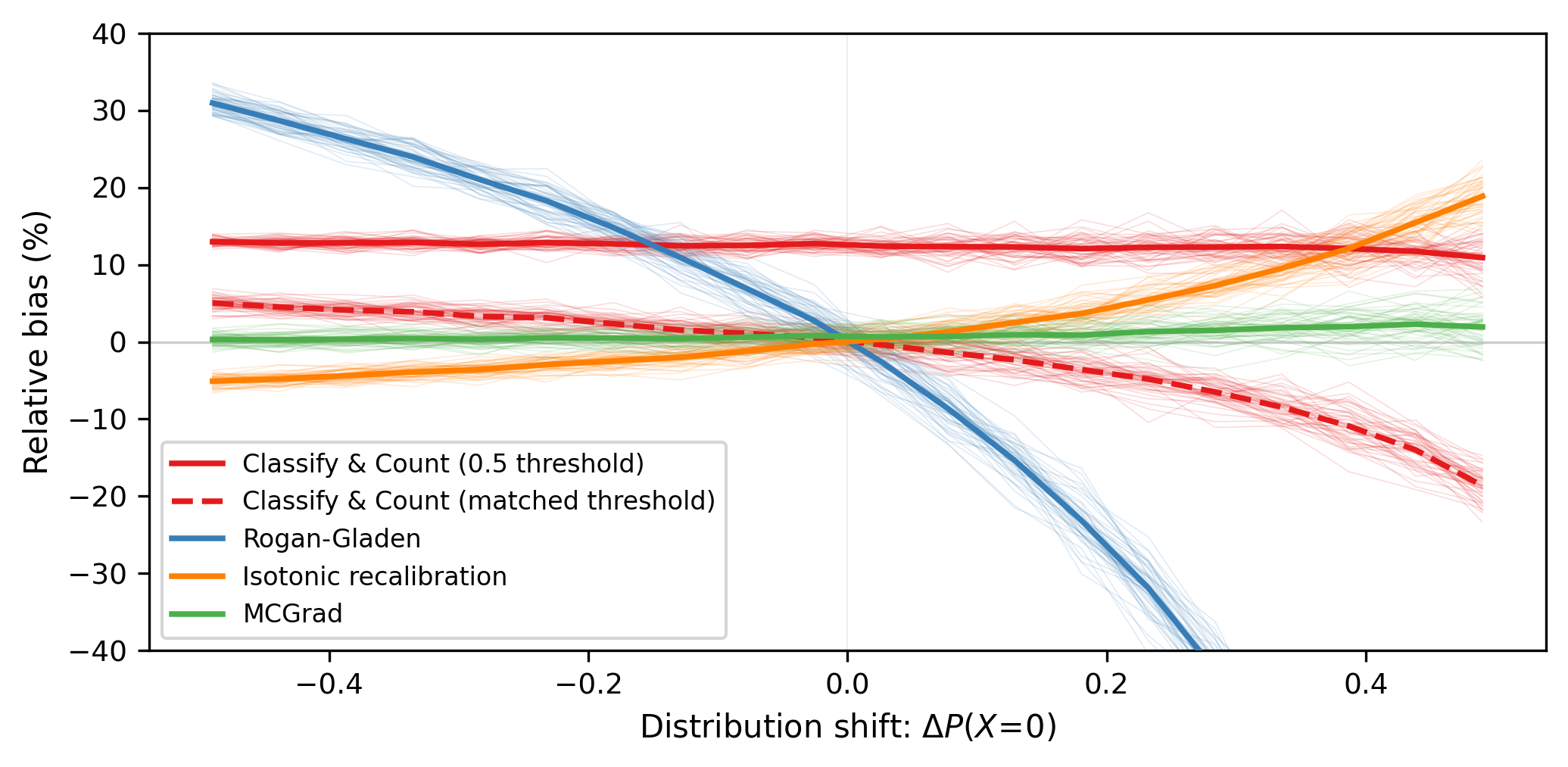}

\emph{Figure 1. Relative prevalence bias under covariate shift, against
the change in \(P(X=0)\) from its calibration value of 0.5. Faint lines
are individual runs; thick lines average over 50 runs. The figure is
cropped at \(\pm40\%\); SI Sections S4--S5 show the full range, all
methods, and RMSE.}

At the calibration distribution, Classify \& Count with the default 0.5
threshold overstates prevalence by about 12\%: good discrimination does
not make counts accurate. The labeled sample can remove this bias in
three common ways: choosing the threshold so that the count reproduces
the calibration prevalence, adjusting the count for the estimated error
rates (Rogan-Gladen), or recalibrating the scores (isotonic regression).
All three are unbiased at the calibration distribution, but each fits a
single correction to the calibration mix of \(X\) and fails when that
mix changes. The matched-threshold count drifts to about \(-19\%\), and
Rogan-Gladen, whose ratio form amplifies changes in the error rates,
exceeds \(\pm30\%\). Isotonic recalibration shows the mechanism most
clearly: it maps a score to the same probability regardless of \(X\),
which is too high for \(X=0\) observations and too low for \(X=1\).
These errors cancel only at the calibration mix, so the bias grows to
about \(+19\%\) when nearly all observations have \(X=0\). MCGrad uses
the same labeled sample and scores but calibrates within each value of
\(X\), so its average bias stays within about 2\% throughout. The
0.5-threshold count is wrong everywhere; its bias barely moves only
because its relative error happens to be similar in the two groups.

\section{Applications}\label{applications}

We apply the method in two settings with known ground truth. In the
first, Claude Opus
4.6\footnote{An open-weights replication with Llama 3.3 70B (SI Section S2) reproduces the qualitative pattern and permits full reproduction.}
codes 29,900 Comparative Agendas Project texts (Baumgartner et al. 2006)
from six sub-populations in four countries and languages for whether
they concern Law \& Crime (CAP major topic 12), scored against the
expert codes. The topic's prevalence is 7.9\% at baseline and 6--20\%
across scenarios. The LLM returns a Yes/No label and an elicited
probability (Tian et al. 2023); we multicalibrate the label and give the
probability to methods that require a score. In the second, a logistic
regression predicts employment from 16 sociodemographic features in
American Community Survey (ACS) microdata (details in SI Sections
S2--S3). In both, we construct targets with known covariate shift:
within-support reweightings (of countries and document types; of age
groups) and out-of-support targets (unseen document types; held-out
states). Uncertainty comes from a bootstrap of the calibration and
target samples that holds the classifier fixed (SI Section S1.9).

\includegraphics[width=1\linewidth,height=\textheight,keepaspectratio]{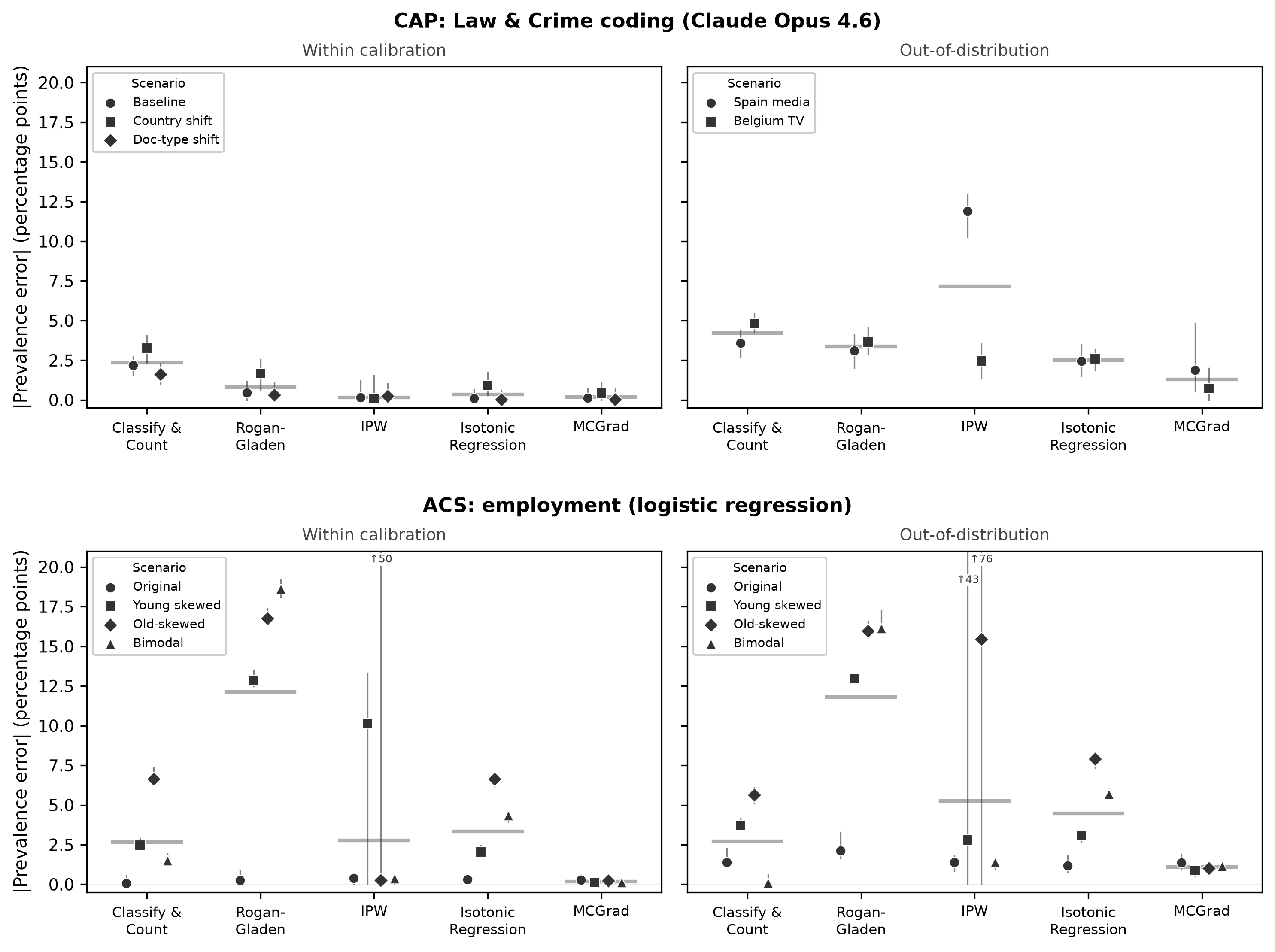}

\emph{Figure 2. Absolute prevalence error (percentage points) by method
and scenario. Top: CAP Law \& Crime coding of 29,900 texts (1947--2018)
with Claude Opus 4.6, multicalibrated on the Yes/No label. Bottom: ACS
employment with logistic regression (2016--2018 microdata; eight source
and six held-out states). Left: within-support targets; right:
out-of-distribution targets. Markers are scenarios, horizontal bars
their means, and vertical lines 95\% bootstrap intervals, with the upper
bound printed where an interval exceeds the axis. Full results in SI
Tables S1--S2.}

Although the LLM discriminates well (per-language AUC about 0.98),
Classify \& Count, the applied default, errs by +1.6 to +4.8pp (Figure
2, top), overstating prevalence by nearly 30\% at baseline. Rogan-Gladen
reduces this error within support but misses by +3.1 to +3.7pp outside
it.\footnote{SI Tables S1--S2 report additional methods, including SLD [@saerens2002adjusting] and PACC. ReadMe [@hopkinsking2010nonparametric], the best-known political-science quantifier, is classifier-free and fails as the label-shift diagnosis predicts, exceeding +60pp under some shifts (SI Section S7).}
Isotonic recalibration is accurate within support (\(\le 0.9\)pp) but
drifts outside it (up to 2.6pp). IPW, with a gradient-boosted propensity
model on the multicalibration features, is near zero within support but
misses by \(-11.9\)pp on Spanish media, whose feature values are absent
from the source. Multicalibration errs by at most 0.4pp within support
and by 0.7 and 1.9pp outside it, with 95\% bootstrap intervals within
5pp (SI Table S2).

The survey application reproduces this ordering with a wider spread
(Figure 2, bottom). Rogan-Gladen fails by 12--19pp under the age shifts,
and Classify \& Count and isotonic regression reach 6--8pp under the
largest. IPW, whose propensity model uses MCGrad's untuned learner, is
accurate on some targets but unstable on all four age-skewed ones: its
point errors reach \(-10.1\) and \(-15.5\)pp, and it errs by more than
5pp in a quarter to three quarters of bootstrap draws (SI Table S1). The
calibration sample is thirty times larger than each target, so target
propensities are small and a few source observations receive extreme
weights (SI Section S3). Multicalibration errs by at most 0.27pp on
within-support targets and by 0.88--1.35pp on held-out states. The
contrast with CAP shows when global recalibration suffices: when the
predictor's errors are roughly homogeneous across the groups a shift
reweights, as they approximately are in CAP. The ACS shifts move age,
which drives employment (76\% employed at ages 25--54, 17\% at 65+), so
group errors are large and heterogeneous, and only feature-conditional
calibration removes them.

\section{Discussion}\label{discussion}

Discriminative validation shows that a classifier separates classes, not
that its prevalence estimates are unbiased. Counting the labels of an
LLM with a per-language AUC of about 0.98 overstated prevalence by
18--44\% of the true rate. Multicalibration removes this bias in the
validation population and keeps it removed as the target shifts along
the calibrated features. Unlike importance weighting, it is fitted once,
before any target is seen, and yields a single instrument for a time
series or a set of sub-corpora without target labels or per-target
re-estimation. Because it removes error within every calibrated group,
not only on average, it also makes group-level estimates comparable:
differences in estimated prevalence across countries, languages or
sources are not just artifacts of a classifier that errs differently in
each. It also leaves the usual workflow intact: prevalence is still the
average of the classifier's outputs, after a one-time adjustment on the
labeled sample already used for validation. Label-shift corrections
remain appropriate when only the frequency of the category changes, not
what its instances look like; when composition changes along observed
features, as across countries, sources or periods, multicalibration is
the reusable choice.

The guarantee has limits. It holds only along calibrated features, so
the calibration data must span the anticipated variation; beyond them
error grows, if only modestly in our out-of-support targets. Calibrating
on more features widens the guarantee but requires enough labels in each
group; in zero-shot settings this reintroduces some hand-coding, though
far less than per-document labeling and only once. Finally, where the
target combines covariate shift with changes in \(P(Y\mid X)\), the
guarantee holds only approximately.

\section*{Funding}\label{funding}
\addcontentsline{toc}{section}{Funding}

This research was conducted as part of the authors' employment at Meta
Platforms, Inc.~and received no specific external funding.

\section*{Acknowledgements}\label{acknowledgements}
\addcontentsline{toc}{section}{Acknowledgements}

The authors used Claude Opus 4.6, 4.7, 5, 5.5, and Gemini 3 Pro to
assist with literature search, code, data analysis, visualization, and
drafting and editing of the manuscript. The authors are entirely
responsible for the scientific content of the paper, which adheres to
the journal's authorship policy.

\section*{Data Availability
Statement}\label{data-availability-statement}
\addcontentsline{toc}{section}{Data Availability Statement}

Replication code and data will be deposited in the Political Analysis
Dataverse and are available at
\url{https://github.com/facebookresearch/multicalibrated_llm_measurement}.
The analyses use Comparative Agendas Project data from Denmark
(Green-Pedersen and Mortensen 2019), Spain (Chaqués-Bonafont et al.
2019, n.d.), the United States (Wilkerson et al. 2025) and Belgium
(Walgrave et al., n.d.), coded under the CAP master codebook (Jones et
al. 2025), and American Community Survey microdata (U.S. Census Bureau
2019), accessed through the folktables package (Ding et al. 2021).

\section*{Competing Interests}\label{competing-interests}
\addcontentsline{toc}{section}{Competing Interests}

All authors are employees of Meta Platforms, Inc.; Milan Vojnovic is
also affiliated with the London School of Economics and Political
Science. Llama 3.3, used in the replication in SI Section S2, is
developed by Meta. Niek Tax, Lorenzo Perini, Fridolin Linder, Daniel
Haimovich, and Milan Vojnovic are authors of MCGrad (Tax et al. 2026),
the open-source multicalibration method evaluated here. The authors
declare no other competing interests.

\section*{References}\label{references}
\addcontentsline{toc}{section}{References}

\protect\phantomsection\label{refs}

\clearpage
\setcounter{secnumdepth}{-2}

\section{SI Appendix}\label{si-appendix}

\subsection{S1. Formal Definitions of Prevalence Estimation
Methods}\label{s1.-formal-definitions-of-prevalence-estimation-methods}

This section defines the prevalence estimation methods used in the
simulation and the applications. The simulation compares seven of them
(Figure S2); main-text Figure 1 shows four of them, with Classify \&
Count at two thresholds.

\textbf{Setup.} Let \(h(X) \in [0,1]\) denote the classifier's
probabilistic prediction for input \(X\), with true label
\(Y \in \{0,1\}\). The goal is to estimate the target prevalence
\(\pi^* = P^*(Y=1)\) using only unlabeled target data
\(\{X_i^*\}_{i=1}^n\) and calibration parameters estimated from a
labeled source dataset.

\subsubsection{S1.1 Uncalibrated
Averaging}\label{s1.1-uncalibrated-averaging}

\[\hat{\pi}_{\text{raw}} = \frac{1}{n} \sum_{i=1}^n h(X_i^*)\]

\subsubsection{S1.2 Classify \& Count}\label{s1.2-classify-count}

Given a threshold \(\tau\) chosen on calibration data:
\[\hat{\pi}_{\text{CC}} = \frac{1}{n} \sum_{i=1}^n \mathbf{1}[h(X_i^*) \geq \tau]\]

In the simulation, we report Classify \& Count at two thresholds:
\(\tau=0.5\), the default decision rule, and the score quantile at which
\(\hat{\pi}_{\text{CC}}\) reproduces the prevalence of the calibration
sample. The second is itself a correction fitted to the labeled sample.
In the ACS application, \(\tau\) is this prevalence-matched threshold;
in the Llama replication (Section S2), \(\tau\) maximizes Youden's \(J\)
on the calibration set (\(\tau\approx0.80\)). For Claude Opus, Classify
\& Count uses the Yes/No label directly.

\subsubsection{S1.3 Rogan-Gladen (Adjusted
Count)}\label{s1.3-rogan-gladen-adjusted-count}

\[\hat{\pi}_{\text{RG}} = \frac{\hat{\pi}_{\text{CC}} - \widehat{\text{FPR}}}{\widehat{\text{TPR}} - \widehat{\text{FPR}}}\]

where \(\widehat{\text{TPR}}\) and \(\widehat{\text{FPR}}\) are
estimated from calibration data at threshold \(\tau\) (Rogan and Gladen
1978). In the simulation, \(\tau=0.5\); in the applications, \(\tau\) is
as for Classify \& Count (Section S1.2).

\subsubsection{S1.4 Probabilistic Adjusted Classify \& Count
(PACC)}\label{s1.4-probabilistic-adjusted-classify-count-pacc}

\[\hat{\pi}_{\text{PACC}} = \frac{\bar{h} - \hat{\mu}_0}{\hat{\mu}_1 - \hat{\mu}_0}\]

where \(\bar{h} = \frac{1}{n}\sum_i h(X_i^*)\),
\(\hat{\mu}_1 = \hat{\mathbb{E}}[h(X)|Y=1]\), and
\(\hat{\mu}_0 = \hat{\mathbb{E}}[h(X)|Y=0]\) are estimated from
calibration data ({González et al.} 2017).

\subsubsection{S1.5 SLD (EMQ)}\label{s1.5-sld-emq}

The Saerens-Latinne-Decaestecker algorithm iterates:

\begin{enumerate}
\def\labelenumi{\arabic{enumi}.}
\tightlist
\item
  Initialize \(\hat{\pi}^{(0)}\) from the source prevalence.
\item
  E-step: Adjust posteriors for the new prior:
  \[\tilde{h}_i^{(t)} = \frac{(\hat{\pi}^{(t)} / \pi_s) \cdot h(X_i^*)}{(\hat{\pi}^{(t)} / \pi_s) \cdot h(X_i^*) + ((1-\hat{\pi}^{(t)}) / (1-\pi_s)) \cdot (1 - h(X_i^*))}\]
\item
  M-step: \(\hat{\pi}^{(t+1)} = \frac{1}{n}\sum_i \tilde{h}_i^{(t)}\)
\item
  Repeat until convergence (Saerens et al. 2002).
\end{enumerate}

SLD requires \(h\) to be a calibrated posterior in the source
population. We therefore run it on the isotonic-recalibrated scores
\(m(h(X))\) of S1.6, not on the raw scores, in the simulation and in all
applications; on raw scores its bias would conflate miscalibration with
the failure of the label-shift assumption.

\subsubsection{S1.6 Global Calibration}\label{s1.6-global-calibration}

Global calibration fits a monotone map \(m\) from scores to
probabilities by isotonic regression of \(Y\) on \(h(X)\) in the
calibration data, and estimates
\(\hat{\pi}_{\text{iso}} = \frac{1}{n}\sum_i m(h(X_i^*))\). The same
estimator is used in the simulation and in both applications. Because
\(m\) depends on the score alone, it is calibrated at the calibration
sample's feature mix but not within feature-defined groups whose scores
overlap.

\subsubsection{S1.7 Multicalibration}\label{s1.7-multicalibration}

In the simulation and in both empirical applications, we use MCGrad (Tax
et al. 2026), a multicalibration algorithm based on gradient boosting.
MCGrad operates in logit space: given a base predictor \(f_0(X)\) with
logit \(F_0(X) = \text{logit}(f_0(X))\), it iteratively fits gradient
boosted decision trees (GBDTs) on the residuals between labels and
current predictions. At each round \(t\), a GBDT \(g_t\) is trained with
the current logit predictions as \texttt{init\_score} and with the
feature matrix consisting of the segment features (categorical and
numerical) augmented by the current logit prediction as an additional
input feature. The logit predictor is then updated as
\(F_{t+1}(X) = \alpha_t \cdot (F_t(X) + g_t(X))\), where \(\alpha_t\) is
a single scalar unshrinkage factor estimated by a one-parameter logistic
regression of the labels on the combined logit \(F_t(X) + g_t(X)\),
applied to the full running logit rather than to the increment alone.
Because \(\alpha_t\) is fit to maximize fit of the combined logit to the
labels at each round, it counteracts the shrinkage induced by the GBDT
learning rate without disturbing the relative structure the trees
discovered; Tax et al. (2026) give conditions under which the resulting
sequence reduces multicalibration error. By including the prediction as
a feature, GBDT splits naturally discover miscalibrated regions in the
joint space of features and score levels, thereby approximating
multicalibration without requiring explicit group specification. The
number of rounds is chosen by cross-validated early stopping (five folds
by default) to prevent overfitting. MCGrad uses LightGBM as the GBDT
implementation. See Tax et al. (2026) for theoretical results and
deployment details. Our claims are algorithm-agnostic: any procedure
producing a multi-accurate, a fortiori multicalibrated, predictor over
the feature class delivers the same guarantee, and several exist
(Hébert-Johnson et al. 2018; Gopalan et al. 2022; Detommaso et al.
2024). An open-source implementation of MCGrad is available at
\url{https://mcgrad.dev}.

\subsubsection{S1.8 Multi-accuracy versus
multicalibration}\label{s1.8-multi-accuracy-versus-multicalibration}

The main text states the universal-adaptability guarantee for a
multicalibrated predictor and computes one with MCGrad. The property
strictly required for prevalence estimation is weaker. Let
\(e(X,Y)=f(X)-Y\). A predictor is \emph{multi-accurate} with respect to
a group class \(\mathcal{G}\) if
\[\mathbb{E}[\mathbf{1}\{X\in G\}e(X,Y)]=0 \qquad \text{for every }G\in\mathcal{G}.\]
This is equivalent to zero mean signed error within every group of
positive probability. Multicalibration implies this condition by
averaging \(\mathbb{E}[Y\mid f(X)=v,X\in G]=v\) over the prediction
values \(v\) within each group.

Let \(P\) and \(P^*\) denote the source and target populations, and let
\(r(X)=dP_X^*/dP_X\) be their density ratio. Under covariate shift and
overlap, \[\mathbb{E}^*[f(X)-Y]=\mathbb{E}[r(X)e(X,Y)].\] Suppose the
target shift is representable by the group class, meaning that for some
\(G_1,\ldots,G_J\in\mathcal{G}\) and coefficients \(a_1,\ldots,a_J\),
\[r(X)=\sum_{j=1}^J a_j\mathbf{1}\{X\in G_j\}.\] Multi-accuracy then
gives \[
\mathbb{E}^*[f(X)-Y]
=\sum_{j=1}^J a_j\mathbb{E}[\mathbf{1}\{X\in G_j\}e(X,Y)]
=0.
\] Thus one source-fitted predictor is unbiased for every target whose
density ratio lies in the linear span of the calibrated group indicators
(Kim et al. 2022). For a single fixed target, only the weaker condition
\(\mathbb{E}[r(X)e(X,Y)]=0\) is necessary.

The same argument gives an approximate result. If a function \(g\) in
the span of the group indicators approximates \(r\), then \[
\left|\mathbb{E}^*[e(X,Y)]\right|
\leq \left|\mathbb{E}[g(X)e(X,Y)]\right|
+\mathbb{E}[|r(X)-g(X)|],
\] where the final term uses \(|e(X,Y)|\leq1\). The first term reflects
remaining multi-accuracy error and the second reflects how well the
calibrated group class represents the target shift.

Multicalibration remains the more useful practical target. It also
controls calibration within score levels and therefore applies when
selection or downstream analysis depends on the model's own scores.
MCGrad (Tax et al. 2026) approximates this property using the feature
and score interactions learned by gradient-boosted trees (Section S1.7).
Other procedures can provide the same population guarantee if they
achieve the required group-level residual balance (Hébert-Johnson et al.
2018; Gopalan et al. 2022; Detommaso et al. 2024).

\subsubsection{S1.9 Estimation and implementation
details}\label{s1.9-estimation-and-implementation-details}

\textbf{Simulation.} The data-generating process is given in the main
text: \(X\sim\text{Bernoulli}(1-P(X=0))\), \(U\sim N(0,1)\) independent
of \(X\), \(P(Y=1\mid X,U)=\sigma(a_X+bU)\) with \(a_0=-2\),
\(a_1=1.5\), \(b=1.5\), and classifier score
\(h=\sigma(a_X+bU+\delta_X)\) with \(\delta_0=0.8\), \(\delta_1=0\).
Each of the 50 runs draws a fresh calibration sample of \(n=10{,}000\)
at \(P(X=0)=0.5\), fits every estimator once, and evaluates bias and
RMSE on 20 fresh unlabeled targets of \(n=10{,}000\) with \(P(X=0)\)
evenly spaced in \([0.01,0.99]\). Bias is relative to each target
sample's realized prevalence. MCGrad uses \(X\) as its single
categorical feature with default hyperparameters. Definitions of all
seven methods are in Section S1.

\textbf{Comparative Agendas Project.} The two campaigns (binary Yes/No
and direct probability elicitation) were run separately to avoid
anchoring. Benchmark implementations: Rogan-Gladen uses TPR/FPR
estimated on the calibration set; IPW estimates the
calibration-vs-target propensity with a default LightGBM classifier, the
learner underlying MCGrad, on the same features MCGrad uses (country,
document type, party, decade, and length), cross-fitted over five folds;
isotonic regression is fit on the scores. Per-scenario prevalence error
is in Table S2; the ReadMe comparison is in Section S7 and the Llama 3.3
70B replication in Section S2.

\textbf{Claude Opus 4.6 elicitation.} The CAP sample contains 30,000
documents, 5,000 from each of the six sub-populations. We classified
them with Claude Opus 4.6 (Anthropic, model identifier
\texttt{claude-opus-4-6}) on 2--3 April 2026, accessed through the
Claude Code command-line interface rather than the API. A Claude Code
session running Opus 4.6 split the sample into 300 shards of 100
documents and dispatched each shard to a sub-agent, which inherits the
parent session's model; 50 sub-agents ran concurrently, in six rounds
per campaign. Each sub-agent classified the documents of its shard
within a single context, so classifications within a shard are not
strictly independent. Sampling parameters were Claude Code defaults and
were not set by us. Texts were truncated to their first 4,000
characters. The two campaigns used separate sub-agents with the
following prompts, where \texttt{\{language\}} is the document's
language, \texttt{\{text\}} the document, and \texttt{\{codebook\}} the
CAP master-codebook definition of Law \& Crime (Jones et al. 2025) given
below.

\begin{quote}
\emph{Binary campaign:} The following text is in \{language\}. Does it
primarily fall under the policy topic ``Law and Crime'' as defined by
the Comparative Agendas Project? \{codebook\} Text: \{text\} Answer Yes
or No.
\end{quote}

\begin{quote}
\emph{Probability campaign:} The following text is in \{language\}.
Consider whether it primarily falls under the policy topic ``Law and
Crime'' as defined by the Comparative Agendas Project. \{codebook\}
Text: \{text\} Without first deciding Yes or No, directly estimate the
probability that this text is about Law and Crime. Provide P(Yes) and
P(No) as your true belief probabilities, summing to 1.0. Respond in
exactly this format: P(Yes): {[}0.XX{]} P(No): {[}0.XX{]}
\end{quote}

\begin{quote}
\emph{Codebook definition:} Law and Crime includes: general law, crime,
and family issues; law enforcement agencies including border, customs,
and specialized enforcement agencies; white collar crime, organized
crime, counterfeiting, fraud, cyber-crime, and money laundering; illegal
drug crime and enforcement, criminal penalties for drug crimes, and
international efforts to combat drug trafficking; court administration,
bail, pre-release, fines, and legal representation; prisons, jails, and
parole systems; juvenile crime and justice, and efforts to reduce
juvenile crime and recidivism; child abuse, child pornography, sexual
exploitation of children, and parental kidnapping; family issues,
domestic violence, child welfare, and family law; domestic criminal and
civil codes; crime control, prevention, and impact of crime; and police
and domestic security responses to terrorism.
\end{quote}

The score is the elicited P(Yes). One shard of 100 Belgian newspaper
articles returned no binary answers and is excluded, leaving the 29,900
documents analyzed in the main text. Because the model is proprietary
and was run without a fixed seed, the classifications cannot be
regenerated exactly; we release all model outputs, the prompt templates
and the shard-management script, and the Llama 3.3 70B replication
(Section S2) can be reproduced in full.

\textbf{CAP data sources.} All CAP files were downloaded from
comparativeagendas.net on 2 March 2026: Danish parliamentary questions
(Green-Pedersen and Mortensen 2019), Spanish oral questions (release
19.1) and El País and El Mundo front pages (Chaqués-Bonafont et al.
2019, n.d.), U.S. congressional bills (release 19.3) (Wilkerson et al.
2025), and Belgian television news and \emph{De Standaard} newspaper
stories (Walgrave et al., n.d.). None carries a DOI. As requested by the
data providers: the data in the Danish Policy Agenda Project were
collected by Christoffer Green-Pedersen and Peter Bjerre Mortensen with
support from the Danish Social Science Research Council and the Research
Foundation at Aarhus University. The Spanish data were originally
collected by Laura Chaqués-Bonafont, Anna M. Palau and Luz M. Muñoz,
with the collaboration of graduate students and the financial support of
the Spanish Ministry of Innovation and Science and AGAUR. The Belgian
data were collected by Stefaan Walgrave and his collaborators (Jeroen
Joly, Anne Hardy, Brandon Zicha, Julie Sevenans, and Tobias Van Assche),
with funding from the European Science Foundation (07-ECRP-008), the
Flemish National Science Foundation (G.0117.11N) and the Belgian Federal
Science Policy (IUAP P7/46). Neither the funders nor the original
collectors of the data bear any responsibility for the analysis reported
here.

\textbf{American Community Survey.} Training states are TX, MI, PA, OH,
IL, GA, NC, VA (2016--2018); held-out test states are CA, NY, FL, WA,
AZ, CO, with in-distribution test \(n\approx 920{,}000\). Age-shifted
targets are produced by importance-weighted resampling. Post-hoc
calibration uses isotonic regression and MCGrad with categorical and
numerical features. The 2016--2018 1-Year person files (U.S. Census
Bureau 2019) were downloaded through folktables (Ding et al. 2021) on 14
April 2026.

\textbf{Uncertainty.} Intervals for the CAP and ACS applications come
from a nonparametric bootstrap that holds the classifier fixed (the Opus
outputs; the ACS logistic regression), so they reflect sampling
variability of the labeled calibration data and of the targets,
conditional on the measurement device. Each replicate resamples with
replacement the calibration set (for CAP, within each sub-population),
the in-distribution test pool, and the out-of-distribution targets;
redraws the reweighted scenarios from the resampled pools with a new
seed; refits every estimator, including MCGrad, isotonic regression, the
Rogan-Gladen rates and the IPW propensity models; and records the signed
error against that replicate's realized target prevalence. Calibration
observations drawn more than once enter as a single row weighted by
their count rather than as repeated rows: MCGrad selects its number of
boosting rounds by cross-validation and IPW cross-fits its propensity
model, and repeated rows would fall on both sides of a fold, favoring
overfitted fits. Point estimates are from the original samples. We
report 95\% percentile intervals from 200 replicates for CAP
(\texttt{cap\_analysis/opus/bootstrap\_cap\_opus.py}) and 50 for ACS
(\texttt{acs\_analysis/bootstrap\_acs.py}), where each replicate refits
IPW on 20,000-observation targets against the 644,000-observation
calibration set. Percentile intervals, rather than normal intervals, are
needed because the IPW errors are heavy-tailed on the age-skewed ACS
targets.

\subsection{S2. Robustness: Replication with Open-Weight LLM (Llama 3.3
70B)}\label{s2.-robustness-replication-with-open-weight-llm-llama-3.3-70b}

The main text reports results using Claude Opus 4.6 as the LLM
measurement device. To verify that the findings are not specific to a
particular model, we replicate the CAP analysis using Llama 3.3 70B
Instruct (Meta 2024), a model in the Llama 3 family ({Grattafiori et
al.} 2024) (4-bit NF4 quantized, run on a single A100 80GB GPU). We
report results for verbalized confidence scores; we also describe token
log-probabilities, which we considered and rejected.

\subsubsection{S2.1 Score Extraction
Methods}\label{s2.1-score-extraction-methods}

\textbf{Log-probabilities.} For each document, the model is prompted
with the CAP codebook definition of Law \& Crime and asked to respond
Yes or No.~The score is extracted from next-token log-probabilities:
\(h(X) = P(\text{Yes}) / (P(\text{Yes}) + P(\text{No}))\). This produces
highly bimodal scores: 23\% of the 105,000 documents score at exactly
0.0 or 1.0, and only 6\% fall in the mid-range {[}0.1, 0.9{]}. Such
scores carry little information about uncertainty, so we do not use
them.

\textbf{Verbalized confidence (2-stage).} A two-stage dialogue first
asks the model to classify the document (Yes/No), then asks it to
estimate the probability that its answer is correct, following the
verbalized-confidence approach of Tian et al. (2023), with an
instruction we added to discourage degenerate outputs (``Note: a
probability of exactly 0.0 or 1.0 would mean absolute certainty, which
is rarely warranted''). The score is \(P(\text{correct})\) if the answer
is Yes and \(1 - P(\text{correct})\) if No.~This produces scores with
negligible boundary mass (75\% in {[}0.1, 0.9{]}) but only 11 unique
score values.

\subsubsection{S2.2 Data and
Calibration}\label{s2.2-data-and-calibration}

The Llama analysis uses the full 105,000-document sample (15,000 per
sub-population for Denmark questions, Spain questions, U.S. bills, and
Belgium newspaper; 30,000 for Spanish media; 15,000 for Belgian TV). The
calibration set (\(n \approx 40{,}000\)) is drawn equally from the four
in-distribution sub-populations. MCGrad is calibrated with categorical
features (country, document type, party) and one numerical feature
(decade). Because this analysis draws on the larger 105,000-document
sample rather than the 29,900-document Opus main-text sample, the
expert-coded true prevalences below differ slightly from Table S2 (e.g.,
8.1\% vs.~7.9\% at baseline); these are sampling differences in the gold
standard, not discrepancies in the method.

\subsubsection{S2.3 Results: Verbalized Confidence
Scores}\label{s2.3-results-verbalized-confidence-scores}

Table S3 shows prevalence estimation error using Llama 3.3 70B with
verbalized confidence scores. Unlike the main-text Opus headline, which
multicalibrates the Yes/No label, MCGrad here multicalibrates the
verbalized score.

{\def\LTcaptype{none} 
\begin{longtable}[]{@{}
  >{\raggedright\arraybackslash}p{(\linewidth - 14\tabcolsep) * \real{0.1250}}
  >{\raggedright\arraybackslash}p{(\linewidth - 14\tabcolsep) * \real{0.1250}}
  >{\raggedright\arraybackslash}p{(\linewidth - 14\tabcolsep) * \real{0.1250}}
  >{\raggedright\arraybackslash}p{(\linewidth - 14\tabcolsep) * \real{0.1250}}
  >{\raggedright\arraybackslash}p{(\linewidth - 14\tabcolsep) * \real{0.1250}}
  >{\raggedright\arraybackslash}p{(\linewidth - 14\tabcolsep) * \real{0.1250}}
  >{\raggedright\arraybackslash}p{(\linewidth - 14\tabcolsep) * \real{0.1250}}
  >{\raggedright\arraybackslash}p{(\linewidth - 14\tabcolsep) * \real{0.1250}}@{}}
\toprule\noalign{}
\begin{minipage}[b]{\linewidth}\raggedright
Scenario
\end{minipage} & \begin{minipage}[b]{\linewidth}\raggedright
Shift Type
\end{minipage} & \begin{minipage}[b]{\linewidth}\raggedright
True Prev.
\end{minipage} & \begin{minipage}[b]{\linewidth}\raggedright
CC
\end{minipage} & \begin{minipage}[b]{\linewidth}\raggedright
RG
\end{minipage} & \begin{minipage}[b]{\linewidth}\raggedright
IPW
\end{minipage} & \begin{minipage}[b]{\linewidth}\raggedright
Iso.
\end{minipage} & \begin{minipage}[b]{\linewidth}\raggedright
MCGrad
\end{minipage} \\
\midrule\noalign{}
\endhead
\bottomrule\noalign{}
\endlastfoot
Baseline & None & 8.1\% & +14.7 & +0.6 & +0.1 & +0.2 & +0.2 \\
Country shift & Within-cal. & 8.7\% & +15.6 & +2.2 & -0.2 & +1.3 &
+0.1 \\
Doc-type shift & Within-cal. & 6.5\% & +16.0 & +1.7 & +0.0 & +0.7 &
+0.1 \\
Spain media & OOD doc type & 19.3\% & +15.4 & +6.6 & -11.1 & -7.2 &
-4.9 \\
Belgium TV & OOD doc type & 11.1\% & +13.3 & -0.0 & -2.6 & -1.6 &
-3.4 \\
\end{longtable}
}

\emph{Table S3: Prevalence estimation error (pp) for Law \& Crime topic
using Llama 3.3 70B with verbalized confidence scores. CC = Classify \&
Count, RG = Rogan-Gladen, IPW = importance-weighted estimation
(cross-fitted gradient-boosted propensity model on country, document
type, party, and decade), Iso. = isotonic regression.}

The pattern is consistent with the main text's Claude Opus results:
MCGrad achieves near-zero error within the calibration distribution
(\(\leq 0.2\)pp) and degrades on OOD populations (-3.4 to -4.9pp).
Several differences are notable:

\begin{itemize}
\tightlist
\item
  \textbf{Higher raw CC error} (+13 to +16pp vs.~+2 to +5pp with Opus).
\item
  \textbf{Comparable MCGrad within-calibration performance}
  (\(\leq 0.2\)pp for Llama, \(\leq 0.4\)pp for Opus), confirming that
  multicalibration corrects for model-specific calibration errors.
\item
  \textbf{Larger OOD error} on Spanish media (-4.9pp vs.~Opus's -1.9pp
  on binary labels / -4.5pp on probability scores), reflecting the
  combination of a weaker base model with a coarser score distribution.
  Note that this comparison mixes elicitation modes: the Llama numbers
  here use verbalized confidence (11 unique values), whereas the Opus
  main-text headline uses binary labels; the closest like-for-like
  comparison is Llama verbalized against Opus's probability-score
  condition (43 unique values, SI Figure S3), and on that comparison the
  gap is consistent with the finer Opus score distribution. The broader
  point (that MCGrad's within-calibration performance is near-identical
  across models and elicitation modes while the input score distribution
  matters chiefly out of support) holds in every comparison.
\end{itemize}

\subsubsection{S2.4 Additional Baseline: SLD on Llama
Scores}\label{s2.4-additional-baseline-sld-on-llama-scores}

The SLD (EMQ) algorithm, designed for label shift rather than covariate
shift, is run on the isotonic-recalibrated Llama scores (S1.5). It is
accurate at baseline (+0.6pp) but errs under the within-calibration
shifts (+5.4pp for the country shift, \(-1.4\)pp for the doc-type shift)
and out of support (+3.0pp on Spanish media, +1.4pp on Belgian TV).

\subsection{S3. Empirical results: ACS employment benchmark and detailed
tables}\label{s3.-empirical-results-acs-employment-benchmark-and-detailed-tables}

As a check with exact ground truth, we estimate employment prevalence
from American Community Survey microdata via the \emph{folktables}
package. The true rate in any subpopulation is known, and the classifier
is an ordinary logistic regression rather than an LLM, so this confirms
the correction is not specific to language models or to noisy gold
labels. We predict employment from 16 sociodemographic features,
training on eight states (2016--2018; approximately 1.5M observations)
and calibrating on a held-out set (\(n\approx 644{,}000\)). Because
employment rates vary sharply by age (76\% for ages 25--54 versus 17\%
for 65+), we construct covariate shifts by resampling the test set to be
young-skewed, old-skewed, or bimodal, yielding true employment rates
from 12.8\% to 46.0\%; all calibration parameters are fixed across
scenarios. We evaluate both on in-distribution states and on six
held-out states.

\includegraphics[width=1\linewidth,height=\textheight,keepaspectratio]{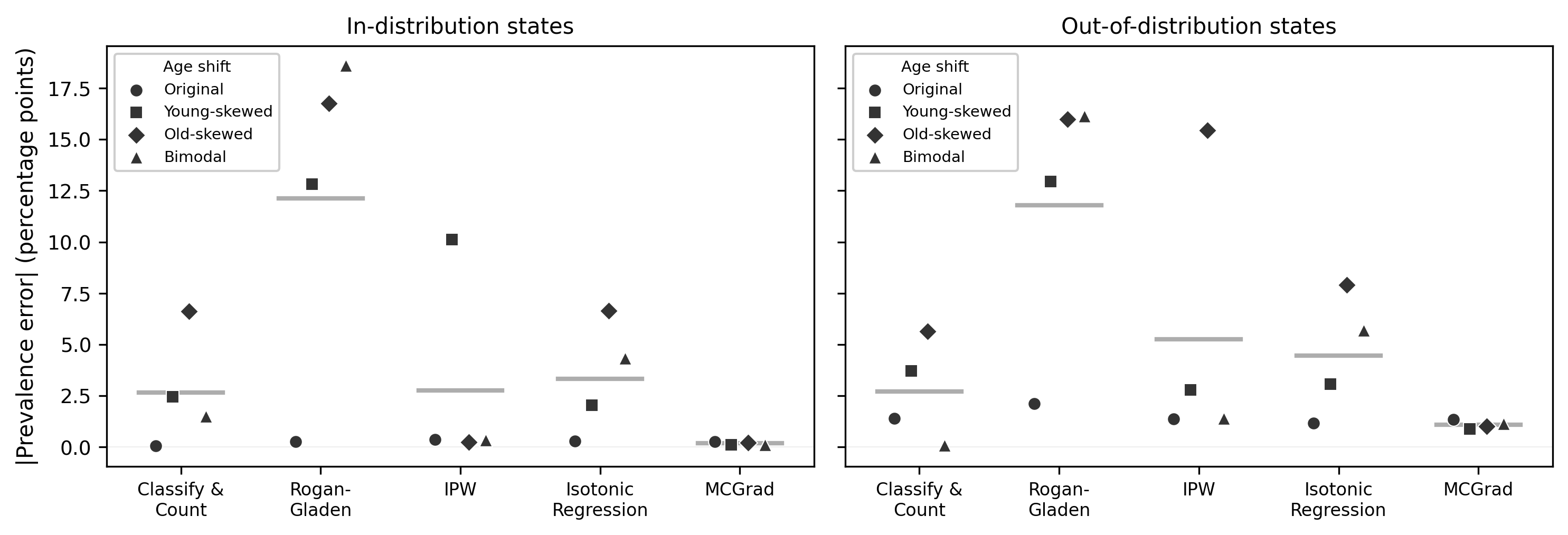}

\emph{Figure S4. Absolute prevalence error (percentage points) for the
ACS employment benchmark, by method and age-shift scenario, for
in-distribution (left) and out-of-distribution (right) states. Marker
shape denotes the synthetic age distribution; horizontal lines are
per-method means. MCGrad is near-exact across all in-distribution
scenarios and degrades only modestly out of distribution; Rogan-Gladen
and isotonic regression grow with the age shift, and IPW collapses on
two targets where its weights degenerate.}

The pattern matches the simulation and CAP results (Figure S4; full
numbers in Table S1). Rogan-Gladen fails by 12--19pp and SLD comparably;
Classify \& Count and isotonic regression show moderate but growing
error (up to 8pp); IPW is accurate on six of eight targets
(\(\le 0.4\)pp in-distribution, \(\le 2.8\)pp out-of-distribution) but
collapses on the in-distribution young-skewed (\(-10.1\)pp) and
out-of-distribution old-skewed (\(-15.5\)pp) targets. Multicalibration
achieves \(\le 0.27\)pp error across all in-distribution scenarios,
including the bimodal shift, and degrades only modestly out of
distribution (0.88--1.35pp), reflecting geographic shift along a
dimension the calibration set did not span. Across both applications the
story is consistent: multicalibration is near-exact when the target's
features lie within the calibration support and degrades predictably
when they do not: the scope condition has visible, interpretable bite
rather than silent failure.

\textbf{Why IPW collapses.} IPW and MCGrad use the same learner
(gradient-boosted trees, default settings, no tuning). For IPW it
estimates the propensity that an observation belongs to the target
rather than the calibration set; the weights are the odds \(p/(1-p)\),
cross-fitted over five folds. The calibration set
(\(n\approx 644{,}000\)) is about thirty times larger than each target
(\(n=20{,}000\)), so the propensities are small on average, and a few
calibration observations placed in near-pure target leaves receive odds
thousands of times the mean. On the two failing targets the effective
sample size of the weights, \((\sum w)^2/\sum w^2\), falls to 518
(young-skewed) and 2 (out-of-distribution old-skewed), and the estimate
is effectively the label of a handful of respondents. IPW can therefore
match multicalibration within support, but only with per-target
monitoring of the weights and tuning when they degenerate. MCGrad needs
neither, because it adjusts predictions bounded in \([0,1]\) and
averages them over every target observation, so no small set of
observations can dominate the estimate, and its boosting rounds are
chosen by cross-validated early stopping on the calibration data.

\subsubsection{Table S1: ACS Employment Prevalence Estimation
Error}\label{table-s1-acs-employment-prevalence-estimation-error}

\begin{center}\resizebox{\linewidth}{!}{%
\begin{tabular}{llrrrrrrrrr}
\toprule
Setting & Age dist. & True prev. & Raw & CC & RG & PACC & SLD & IPW & Iso. & MCGrad \\
\midrule
In-Dist & Original & 46.0\% & $-0.31$ & $-0.07$ & $+0.26$ & $-0.07$ & $+0.01$ & $-0.38$ & $-0.30$ & $-0.27$ \\
 & & & $[-0.6,\,+0.3]$ & $[-0.5,\,+0.4]$ & $[-0.9,\,+0.8]$ & $[-0.9,\,+0.6]$ & $[-0.9,\,+0.6]$ & $[-0.6,\,+0.3]$ & $[-0.6,\,+0.3]$ & $[-0.6,\,+0.3]$ \\[2pt]
In-Dist & Young-skewed & 12.8\% & $+1.93$ & $+2.47$ & $-12.82$ & $-12.82$ & $-12.82$ & $-10.12$ & $+2.04$ & $-0.11$ \\
 & & & $[+1.7,\,+2.3]$ & $[+2.2,\,+2.9]$ & $[-13.5,\,-12.5]$ & $[-13.5,\,-12.5]$ & $[-13.1,\,-12.2]$ & $[-13.3,\,+8.6]$ & $[+1.8,\,+2.4]$ & $[-0.3,\,+0.2]$ \\[2pt]
In-Dist & Old-skewed & 16.8\% & $+7.23$ & $-6.62$ & $-16.77$ & $-16.77$ & $-16.76$ & $+0.24$ & $+6.65$ & $+0.22$ \\
 & & & $[+6.8,\,+7.5]$ & $[-7.3,\,-6.4]$ & $[-17.4,\,-16.4]$ & $[-17.4,\,-16.4]$ & $[-17.4,\,-16.4]$ & $[-17.1,\,+50.0]$ & $[+6.2,\,+7.0]$ & $[-0.3,\,+0.4]$ \\[2pt]
In-Dist & Bimodal & 21.1\% & $+4.57$ & $-1.50$ & $-18.62$ & $-19.97$ & $-16.79$ & $+0.35$ & $+4.33$ & $+0.12$ \\
 & & & $[+4.2,\,+4.8]$ & $[-1.9,\,-1.2]$ & $[-19.2,\,-18.1]$ & $[-20.5,\,-19.4]$ & $[-17.6,\,-16.4]$ & $[-0.1,\,+0.5]$ & $[+3.9,\,+4.6]$ & $[-0.3,\,+0.3]$ \\[2pt]
OOD & Original & 45.1\% & $+1.15$ & $+1.40$ & $+2.12$ & $+2.13$ & $+2.25$ & $+1.38$ & $+1.17$ & $+1.35$ \\
 & & & $[+0.8,\,+1.8]$ & $[+1.1,\,+2.3]$ & $[+1.6,\,+3.3]$ & $[+1.3,\,+2.9]$ & $[+1.4,\,+3.0]$ & $[+0.8,\,+1.8]$ & $[+0.8,\,+1.8]$ & $[+1.0,\,+1.9]$ \\[2pt]
OOD & Young-skewed & 13.0\% & $+2.93$ & $+3.73$ & $-12.96$ & $-12.96$ & $-12.39$ & $-2.80$ & $+3.08$ & $+0.88$ \\
 & & & $[+2.5,\,+3.1]$ & $[+3.5,\,+4.1]$ & $[-13.3,\,-12.7]$ & $[-13.3,\,-12.7]$ & $[-12.8,\,-11.9]$ & $[-13.2,\,+42.7]$ & $[+2.7,\,+3.2]$ & $[+0.5,\,+1.0]$ \\[2pt]
OOD & Old-skewed & 16.0\% & $+8.47$ & $-5.64$ & $-15.97$ & $-15.97$ & $-15.97$ & $-15.45$ & $+7.91$ & $+1.01$ \\
 & & & $[+7.9,\,+8.7]$ & $[-6.1,\,-5.1]$ & $[-16.6,\,-15.5]$ & $[-16.6,\,-15.5]$ & $[-16.6,\,-15.5]$ & $[-16.2,\,+75.6]$ & $[+7.4,\,+8.2]$ & $[+0.6,\,+1.4]$ \\[2pt]
OOD & Bimodal & 20.8\% & $+5.91$ & $+0.09$ & $-16.14$ & $-17.27$ & $-15.63$ & $+1.39$ & $+5.69$ & $+1.13$ \\
 & & & $[+5.5,\,+6.2]$ & $[-0.6,\,+0.3]$ & $[-17.3,\,-15.9]$ & $[-18.1,\,-16.8]$ & $[-16.4,\,-15.2]$ & $[+1.0,\,+1.6]$ & $[+5.4,\,+6.0]$ & $[+0.9,\,+1.4]$ \\[2pt]
\bottomrule
\end{tabular}}
\end{center}

\emph{Prevalence estimation error in percentage points (pp) under
synthetic age distribution shift. Raw = uncalibrated averaging, CC =
Classify \& Count, RG = Rogan-Gladen, IPW = importance-weighted
prevalence estimation (cross-fitted default LightGBM propensity model on
all 16 features), Iso. = Isotonic regression. Brackets: 95\% percentile
intervals from 50 bootstrap replicates (Section S1.9).}

\subsubsection{Table S2: CAP Law \& Crime Prevalence Estimation Error
(Claude Opus
4.6)}\label{table-s2-cap-law-crime-prevalence-estimation-error-claude-opus-4.6}

\begin{center}\resizebox{\linewidth}{!}{%
\begin{tabular}{llrrrrrrrr}
\toprule
Scenario & Shift type & True prev. & CC & RG & SLD & IPW & Iso. & \shortstack[r]{MC\\(binary)} & \shortstack[r]{MC\\(scores)} \\
\midrule
Baseline & None & 7.9\% & $+2.2$ & $+0.5$ & $+0.1$ & $+0.2$ & $+0.1$ & $+0.1$ & $+0.2$ \\
 & & & $[+1.6,\,+2.8]$ & $[-0.2,\,+1.2]$ & $[-0.5,\,+0.8]$ & $[-1.1,\,+1.2]$ & $[-0.5,\,+0.6]$ & $[-0.5,\,+0.7]$ & $[-0.4,\,+0.6]$ \\[2pt]
Country shift & Within-cal. & 8.4\% & $+3.3$ & $+1.7$ & $+1.4$ & $+0.1$ & $+0.9$ & $+0.4$ & $+0.4$ \\
 & & & $[+2.4,\,+4.0]$ & $[+0.6,\,+2.6]$ & $[+0.8,\,+2.7]$ & $[-1.3,\,+1.5]$ & $[+0.3,\,+1.7]$ & $[-0.5,\,+1.1]$ & $[-0.4,\,+1.0]$ \\[2pt]
Doc-type shift & Within-cal. & 6.3\% & $+1.6$ & $-0.3$ & $-0.6$ & $+0.2$ & $+0.0$ & $-0.0$ & $+0.1$ \\
 & & & $[+1.0,\,+2.3]$ & $[-1.1,\,+0.5]$ & $[-1.2,\,+0.1]$ & $[-1.0,\,+1.0]$ & $[-0.6,\,+0.6]$ & $[-0.7,\,+0.7]$ & $[-0.5,\,+0.7]$ \\[2pt]
Spain media & OOD doc type & 19.5\% & $+3.6$ & $+3.1$ & $+3.8$ & $-11.9$ & $-2.5$ & $-1.9$ & $-4.5$ \\
 & & & $[+2.7,\,+4.4]$ & $[+2.0,\,+4.1]$ & $[+2.4,\,+5.3]$ & $[-13.0,\,-10.2]$ & $[-3.5,\,-1.5]$ & $[-4.8,\,-0.5]$ & $[-6.6,\,-2.6]$ \\[2pt]
Belgium TV & OOD doc type & 11.1\% & $+4.8$ & $+3.7$ & $+4.7$ & $-2.4$ & $+2.6$ & $+0.7$ & $+1.6$ \\
 & & & $[+4.3,\,+5.4]$ & $[+2.9,\,+4.5]$ & $[+3.9,\,+5.4]$ & $[-3.5,\,-1.4]$ & $[+1.8,\,+3.2]$ & $[-0.8,\,+2.0]$ & $[+0.6,\,+2.6]$ \\[2pt]
\bottomrule
\end{tabular}}
\end{center}

\emph{CC = Classify \& Count (fraction of Yes labels); RG = Rogan-Gladen
adjustment on binary labels; SLD = Saerens-Latinne-Decaestecker (label
shift, applied to isotonic-recalibrated probability scores); IPW =
importance-weighted estimation (target-specific density ratio from a
cross-fitted gradient-boosted propensity model); Iso. = isotonic
regression on probability scores; MC (binary) = MCGrad on binary labels
with base-rate initialization; MC (scores) = MCGrad on probability
scores. Brackets: 95\% percentile intervals from 200 bootstrap
replicates (Section S1.9).}

\subsection{S4. Simulation: Design and
RMSE}\label{s4.-simulation-design-and-rmse}

\textbf{Design.} Each observation has a binary feature \(X\in\{0,1\}\)
and a continuous signal \(U\sim N(0,1)\) independent of \(X\). The
outcome follows \(P(Y=1\mid X,U)=\sigma(a_X+1.5\,U)\) with \(a_0=-2\)
and \(a_1=1.5\), so \(P(Y=1\mid X=0)\approx0.19\) and
\(P(Y=1\mid X=1)\approx0.75\); this relationship is the same in every
population, and only \(P(X)\) shifts. The classifier's score is
\(\sigma(a_X+1.5\,U+\delta_X)\) with \(\delta_0=0.8\) and
\(\delta_1=0\): it is calibrated for \(X=1\) observations and overstates
the probability for \(X=0\) observations. Each run draws a labeled
calibration sample of 10,000 observations at \(P(X=0)=0.5\), fits every
estimator once, and applies it to fresh unlabeled targets of 10,000
observations on a grid of 20 values of \(P(X=0)\) between 0.01 and 0.99.
Classify \& Count is evaluated at the default 0.5 threshold and at the
threshold that reproduces the calibration prevalence; Rogan-Gladen uses
0.5 (Section S1). We report averages over 50 runs
(\texttt{simulation/helpers.py}).

\textbf{RMSE.} The main-text simulation (Figure 1) reports bias across
the shift gradient. Here we add the root mean squared error for Classify
\& Count at the 0.5 threshold, Rogan-Gladen, isotonic recalibration, and
MCGrad.

\includegraphics[width=1\linewidth,height=\textheight,keepaspectratio]{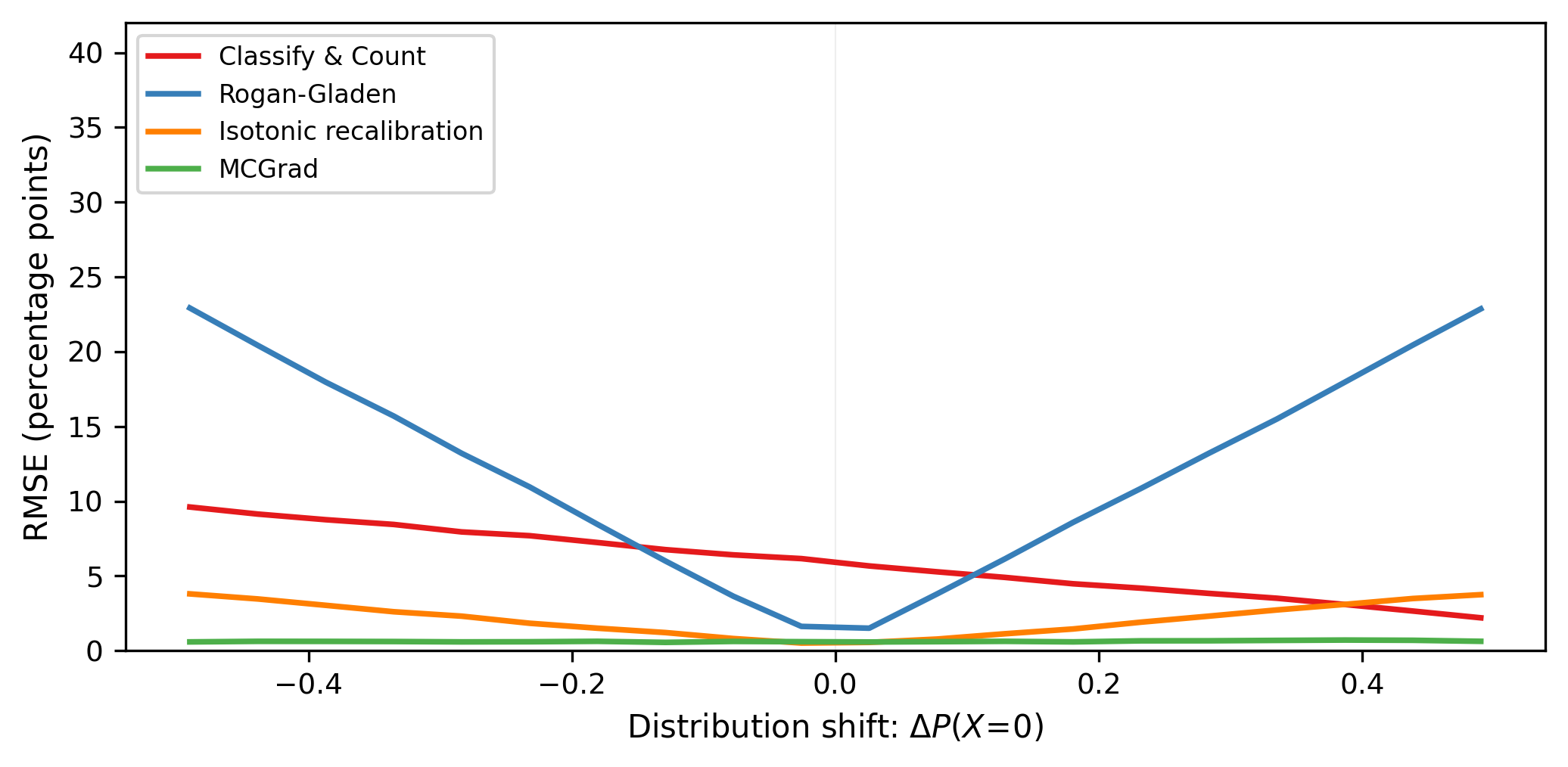}

\emph{Figure S1: Root mean squared error (RMSE) under covariate shift
for Classify \& Count at the 0.5 threshold, Rogan-Gladen, isotonic
recalibration, and MCGrad, averaged over 50 simulation runs. RMSE
closely tracks absolute bias for all methods, confirming that variance
is small relative to bias at this sample size. MCGrad has the lowest
RMSE once the target departs from the calibration distribution.}

\subsection{S5. Simulation: All
Methods}\label{s5.-simulation-all-methods}

\includegraphics[width=1\linewidth,height=\textheight,keepaspectratio]{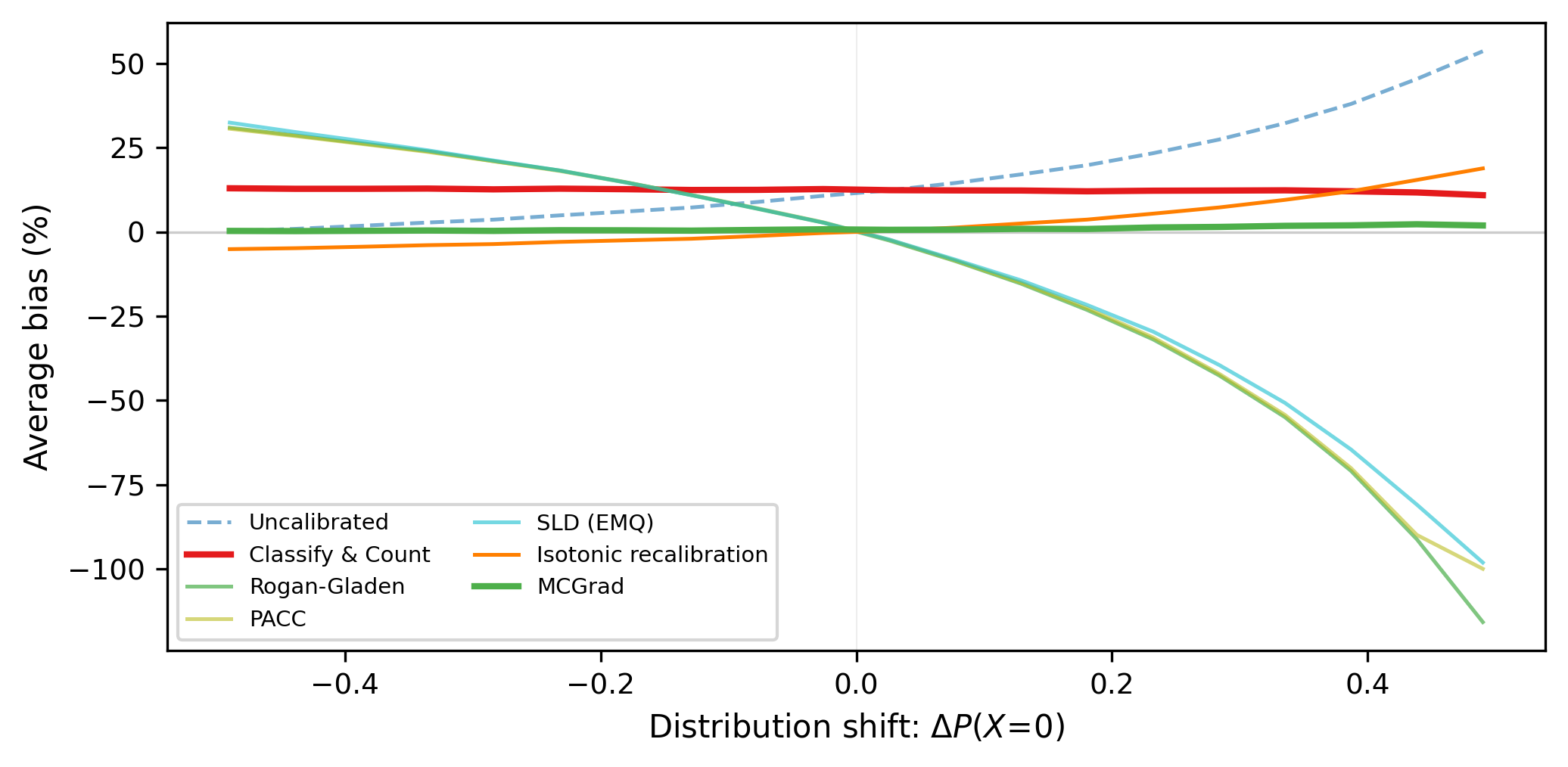}

\emph{Figure S2: Simulation bias curves for all seven methods
(uncalibrated averaging, Classify \& Count, Rogan-Gladen, PACC, SLD/EMQ,
isotonic recalibration, MCGrad). Rogan-Gladen and PACC fail badly under
shift: at the most extreme shift Rogan-Gladen passes \(-100\%\) (a
negative prevalence) and PACC, truncated to \([0,1]\), sits at the
\(-100\%\) floor. SLD, run on isotonic-recalibrated scores, is unbiased
at the calibration distribution but tracks Rogan-Gladen under shift
(about \(+32\%\) and \(-98\%\) at the two extremes), because it
attributes the change in the score distribution to a change in the class
prior. Isotonic recalibration is unbiased at the calibration
distribution and drifts to about \(+19\%\) at the most extreme shift.
MCGrad stays within about 2\% throughout.}

\subsection{S6. Claude Opus Score
Distribution}\label{s6.-claude-opus-score-distribution}

\includegraphics[width=1\linewidth,height=\textheight,keepaspectratio]{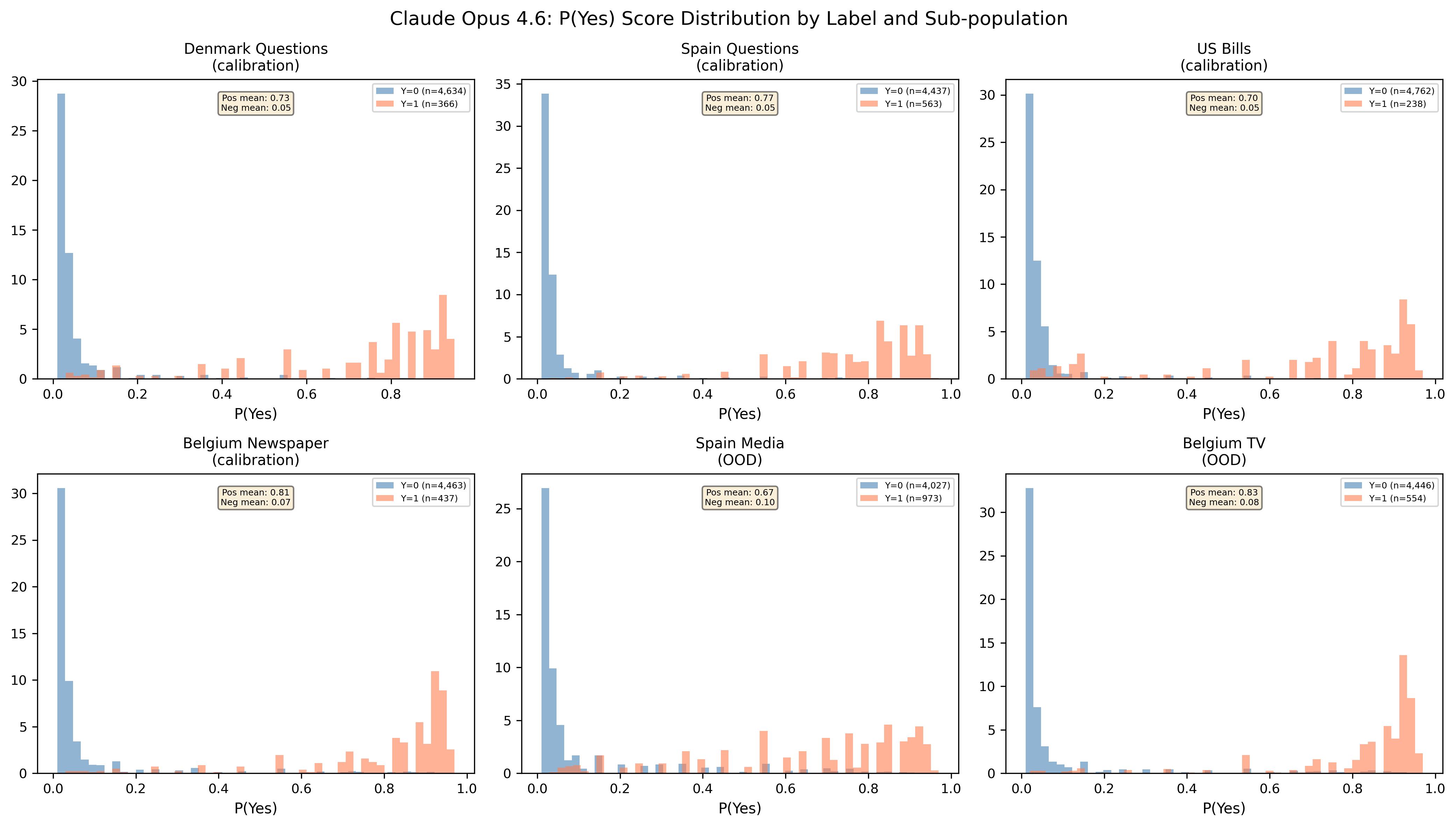}

\emph{Figure S3: Claude Opus 4.6 P(Yes) score distribution by label
across six CAP sub-populations. Scores are well-separated (mean 0.75 for
positives vs.~0.07 for negatives) with 43 unique values and no boundary
mass.}

For reference, the discrimination figures reported in the main text are
tabulated here. Pooled over the four in-distribution sub-populations,
AUC is 0.959 for the binary (Yes/No) condition and 0.987 for the
probability-score condition (0.940 and 0.981 over all six); per-language
AUCs of the probability scores range from 0.978 to 0.985. The
probability-score condition has 43 unique values (this figure); the
verbalized-confidence elicitation used in the Llama replication produces
11 unique values (SI Section S2). All discrimination metrics are
computed against the CAP expert codes as the gold standard.

\subsection{S7. Relationship to the ReadMe Family of
Quantifiers}\label{s7.-relationship-to-the-readme-family-of-quantifiers}

The political-science quantification literature is anchored by ReadMe
(Hopkins and King 2010) and its successor ReadMe2 (Jerzak et al. 2023),
which estimate category proportions directly from document-feature
distributions without per-document classification. Classic ReadMe, like
the Saerens-Latinne-Decaestecker (SLD) algorithm (Saerens et al. 2002)
we report in Tables S1--S2 and many of the quantifiers surveyed by
{González et al.} (2017), assumes \emph{label shift}: the
class-conditional feature distribution \(P(X \mid Y)\) is stable across
the labeled and target sets while the class prior \(P(Y)\) may change.
ReadMe2 relaxes this by choosing feature summaries that are more stable
between labeled and target sets; we do not analyze its identifying
restriction formally here, and like the others it is estimated for a
given labeled/target pair.

The regime studied in the main text is the opposite: \emph{covariate
shift}, where \(P(X)\) changes across populations while \(P(Y\mid X)\)
is stable. Which assumption fits is sometimes argued from causal
direction (Schölkopf et al. 2012). If the category causes the features,
as a topic arguably causes the words used to discuss it, then
\(P(X\mid Y)\) is the stable part and label shift is the natural model.
If the features cause the category, \(P(Y\mid X)\) is stable and
covariate shift is natural. For measurement, what matters more is how
the target differs from the source. When the target is formed by
selecting on features (a different country, venue, period, or document
type, or a corpus that simply contains more of one kind of document),
the chance that a given document belongs to the category does not
change, so \(P(Y\mid X)\) is stable whatever the causal direction. The
within-category feature distribution does change, because the mix of
documents within each category changes: a law-and-crime document from a
corpus heavy in Belgian TV reads differently from one in a corpus of US
bills. The label-shift identity that ReadMe solves is then misspecified.
Label shift is the better model when the target differs only in how
common the category is, for example when more law-and-crime events occur
but each is written about the same way.

We test this directly by benchmarking ReadMe on the CAP corpora, using
the same calibration set and shift scenarios as Table S2. We report two
implementations on the identical documents: a from-scratch
implementation of the Hopkins-King estimator (binary word-presence
features summarized over 300 random feature subsets of size 15;
\texttt{cap\_analysis/readme\_baseline.py}), and the official IQSS
\texttt{readme} package's classic estimator (\texttt{readme0}; 200
subsets, vocabulary 2,500). Both are classifier-free: they use the
document text and the calibration labels, not the LLM's outputs.

{\def\LTcaptype{none} 
\begin{longtable}[]{@{}
  >{\raggedright\arraybackslash}p{(\linewidth - 8\tabcolsep) * \real{0.2000}}
  >{\raggedright\arraybackslash}p{(\linewidth - 8\tabcolsep) * \real{0.2000}}
  >{\raggedright\arraybackslash}p{(\linewidth - 8\tabcolsep) * \real{0.2000}}
  >{\raggedright\arraybackslash}p{(\linewidth - 8\tabcolsep) * \real{0.2000}}
  >{\raggedright\arraybackslash}p{(\linewidth - 8\tabcolsep) * \real{0.2000}}@{}}
\toprule\noalign{}
\begin{minipage}[b]{\linewidth}\raggedright
Scenario
\end{minipage} & \begin{minipage}[b]{\linewidth}\raggedright
Shift
\end{minipage} & \begin{minipage}[b]{\linewidth}\raggedright
True
\end{minipage} & \begin{minipage}[b]{\linewidth}\raggedright
ReadMe (ours)
\end{minipage} & \begin{minipage}[b]{\linewidth}\raggedright
ReadMe (official \texttt{readme0})
\end{minipage} \\
\midrule\noalign{}
\endhead
\bottomrule\noalign{}
\endlastfoot
Baseline & none & 7.9\% & +0.8 & +5.3 \\
Country shift & within-cal. & 8.4\% & +67.6 & +60.1 \\
Doc-type shift & within-cal. & 6.3\% & +7.1 & -5.0 \\
Spain media & OOD doc type & 19.5\% & +4.5 & +44.5 \\
Belgium TV & OOD doc type & 11.1\% & +13.5 & +51.5 \\
\end{longtable}
}

\emph{Table S4: ReadMe prevalence-estimation error (pp) on the CAP
corpora; identical calibration set and scenarios as Table S2. Both
estimators are classifier-free (text only). For comparison, MCGrad
(Table S2) stays within 1.9pp across all scenarios.}

The two implementations differ in magnitude, and on the document-type
shift even in sign, which depends on vocabulary, feature weighting, and
profile-matching choices. ReadMe is near-exact \emph{only} without shift
(our implementation attains +0.8pp at baseline, validating it, while the
official \texttt{readme0} is already +5 to +7pp at baseline on this
short, multilingual corpus) and is inaccurate and highly unstable under
every shift, severely so out of support (the official estimator reaches
+44 to +52pp on the held-out genres; the within-calibration country
shift exceeds +60pp in both implementations). The component estimates
are also unstable: across the 300 random feature subsets, the standard
deviation of our implementation's subset-level estimates is 11pp at
baseline and 32--42pp under shift. This is the dispersion of the
components, not the uncertainty of their average. The mechanism is
interpretable: ReadMe's binary word-presence profiles are sensitive to
document length, so when the target's mean feature density departs from
the calibration set's (long Belgian newspaper documents activate roughly
twice as many features as the calibration average; short Spanish-media
and Belgian-TV snippets far fewer), the within-class feature
distribution shifts and the label-shift identity breaks. This is the
covariate-shift failure the theory predicts, shown directly for the
best-known political-science quantifier rather than inferred from SLD.

ReadMe2 (Jerzak et al. 2023) refines the feature summary (word-vector
``vector summaries'' and a learned matched projection) but still infers
proportions from category-conditional feature summaries estimated on the
labeled set; whether its identifying restriction holds under the
covariate shift studied here has not been established. We installed and
ran the official ReadMe2 estimator; a faithful evaluation requires its
\texttt{undergrad} vector-summary features, and when supplied instead
with multilingual-BERT document embeddings the estimator erred by
roughly +20pp even at the no-shift baseline, mis-specified for that
feature representation and the rare (\textasciitilde8\%) positive class,
so we do not report it as a fair head-to-head comparison. The defensible
empirical statement is the classic-ReadMe result in Table S4.

\subsection{S8. Calibration
Diagnostics}\label{s8.-calibration-diagnostics}

Table S5 reports calibration before and after MCGrad for every analysis,
on held-out data only: the in-distribution test split and, for ACS and
CAP, the out-of-distribution sets. ECCE is the estimated cumulative
calibration error of Arrieta-Ibarra et al. (2022), the range of the
running sum of label-minus-score differences over observations sorted by
score; we report it in percentage points and in units of its standard
deviation under perfect calibration (\(\sigma\)). MCE (Guy et al. 2025)
is the largest ECCE \(\sigma\) over subgroups formed from the features
MCGrad uses, up to three-way intersections (default
\texttt{mcgrad.metrics} settings); its absolute value rescales that
maximum by the global standard deviation. The pre-calibration score is
the classifier or LLM score, except for CAP Opus binary, where MCGrad's
input is the constant calibration base rate and the LLM label enters as
a feature. For that variant the pre-calibration global ECCE only
reflects arbitrary ordering among tied scores (and, out of distribution,
the shift in prevalence), while its MCE reflects the ignored label.

Within the calibration distribution, MCGrad brings global ECCE to
1.0--2.0\(\sigma\) in every analysis, consistent with perfect
calibration, and the worst subgroup to at most 0.71pp. The ACS
in-distribution MCE of 6.2\(\sigma\) corresponds to 0.22pp and is
detectable only because the test set has about 920,000 observations. Out
of distribution, calibration degrades in line with the prevalence error
in Tables S1--S3. Because the running sum ends at the mean
label-minus-score difference, ECCE bounds the absolute error of the
averaged prediction, and the post-calibration ECCE on the
out-of-distribution sets (0.75--5.09pp) is close to the corresponding
MCGrad error. For Opus scores on Spain media and for the ACS held-out
states, MCGrad makes global calibration worse (1.74 to 4.57pp and 1.25
to 1.46pp).

{\def\LTcaptype{none} 
\begin{longtable}[]{@{}
  >{\raggedright\arraybackslash}p{(\linewidth - 10\tabcolsep) * \real{0.1667}}
  >{\raggedright\arraybackslash}p{(\linewidth - 10\tabcolsep) * \real{0.1667}}
  >{\raggedright\arraybackslash}p{(\linewidth - 10\tabcolsep) * \real{0.1667}}
  >{\raggedright\arraybackslash}p{(\linewidth - 10\tabcolsep) * \real{0.1667}}
  >{\raggedright\arraybackslash}p{(\linewidth - 10\tabcolsep) * \real{0.1667}}
  >{\raggedright\arraybackslash}p{(\linewidth - 10\tabcolsep) * \real{0.1667}}@{}}
\toprule\noalign{}
\begin{minipage}[b]{\linewidth}\raggedright
Analysis
\end{minipage} & \begin{minipage}[b]{\linewidth}\raggedright
Evaluation set
\end{minipage} & \begin{minipage}[b]{\linewidth}\raggedright
ECCE (pp)
\end{minipage} & \begin{minipage}[b]{\linewidth}\raggedright
ECCE \(\sigma\)
\end{minipage} & \begin{minipage}[b]{\linewidth}\raggedright
MCE (pp)
\end{minipage} & \begin{minipage}[b]{\linewidth}\raggedright
MCE \(\sigma\)
\end{minipage} \\
\midrule\noalign{}
\endhead
\bottomrule\noalign{}
\endlastfoot
Simulation & In-dist. test & 5.78 → 0.71 & 15.1 → 2.0 & 7.67 → 0.71 &
20.1 → 2.0 \\
ACS & In-dist. test & 0.85 → 0.06 & 22.0 → 1.7 & 3.46 → 0.22 & 89.2 →
6.2 \\
ACS & OOD states & 1.25 → 1.46 & 57.0 → 71.8 & 3.75 → 1.47 & 170.8 →
72.2 \\
Opus, binary & In-dist. test & 0.93 → 0.19 & 2.8 → 1.0 & 21.34 → 0.67 &
63.5 → 3.6 \\
Opus, binary & Spain media & 11.41 → 2.48 & 29.6 → 8.0 & 30.84 → 8.34 &
80.1 → 27.0 \\
Opus, binary & Belgium TV & 3.08 → 0.75 & 8.0 → 2.8 & 23.68 → 0.93 &
61.5 → 3.4 \\
Opus, scores & In-dist. test & 3.34 → 0.30 & 12.4 → 1.7 & 3.34 → 0.57 &
12.4 → 3.3 \\
Opus, scores & Spain media & 1.74 → 4.57 & 4.4 → 14.3 & 1.91 → 4.57 &
4.9 → 14.3 \\
Opus, scores & Belgium TV & 5.45 → 1.78 & 17.2 → 7.3 & 5.45 → 1.78 &
17.2 → 7.3 \\
Llama, scores & In-dist. test & 18.10 → 0.22 & 88.0 → 1.4 & 18.10 → 0.46
& 88.0 → 3.0 \\
Llama, scores & Spain media & 16.17 → 5.09 & 93.7 → 32.2 & 16.17 → 5.09
& 93.7 → 32.2 \\
Llama, scores & Belgium TV & 14.04 → 3.44 & 68.8 → 18.4 & 14.04 → 3.44 &
68.8 → 18.4 \\
\end{longtable}
}

\emph{Table S5: Calibration before → after MCGrad on held-out data; Opus
and Llama rows are the CAP analyses. ECCE = estimated cumulative
calibration error (Arrieta-Ibarra et al. 2022); MCE = multicalibration
error, the maximum ECCE over feature-defined subgroups (Guy et al.
2025); pp = percentage points; \(\sigma\) = standard deviations under
perfect calibration. Simulation: one calibration and one evaluation
sample of 10,000 at \(P(X=0)=0.5\), subgroups on \(X\). ACS subgroups on
all 16 features; CAP subgroups on country, document type, party, decade,
and (Opus) text length, plus the LLM label for the binary variant.
Produced by the run scripts
(\texttt{paper/images/*\_calibration.json}).}

\end{document}